\documentclass[manuscript]{acmart}

\AtBeginDocument{%
  \providecommand\BibTeX{{%
    \normalfont B\kern-0.5em{\scshape i\kern-0.25em b}\kern-0.8em\TeX}}}

\usepackage{mathtools}
\usepackage{amsthm}
\usepackage{microtype}
\usepackage{subfigure}
\usepackage{algorithm}
\usepackage{algorithmic}
\usepackage{mathtools}

\theoremstyle{plain}
\newtheorem{theorem}{Theorem}[section]

\theoremstyle{definition}
\newtheorem{definition}[theorem]{Definition}

\theoremstyle{remark}

\newcommand\given[1][]{\:#1\vert\:}

\begin{document}

\title{Multiscale Graph Comparison via the Embedded Laplacian Discrepancy}

\author{Edric Tam}
\affiliation{%
  \institution{Duke University}
  \streetaddress{Department of Statistical Science}
  \city{Durham}
  \state{North Carolina}
  \country{USA}
  \postcode{27705}
}
\email{edric.tam@duke.edu}
\author{David Dunson}
\affiliation{%
  \institution{Duke University}
  \streetaddress{Department of Statistical Science}
  \city{Durham}
  \state{North Carolina}
  \country{USA}
  \postcode{27705}
}
\email{dunson@duke.edu}

\renewcommand{\shortauthors}{Tam and Dunson.}

\begin{abstract}
Laplacian eigenvectors capture natural community structures on graphs and are widely used in spectral clustering and manifold learning. The use of Laplacian eigenvectors as embeddings for the purpose of multiscale graph comparison has however been limited. Here we propose the Embedded Laplacian Discrepancy (ELD) as a simple and fast approach to compare graphs (of potentially different sizes) based on the similarity of the graphs' community structures. The ELD operates by representing graphs as point clouds in a common, low-dimensional space, on which a natural Wasserstein-based distance can be efficiently computed. A main challenge in comparing graphs through any eigenvector-based approaches is the potential ambiguity that could arise due to sign-flips and basis symmetries. The ELD leverages a simple symmetrization trick to bypass any sign ambiguities. For comparing graphs that do not have any ambiguities due to basis symmetries (i.e. the spectrums are simple), we show that the ELD becomes a natural pseudo-metric that enjoys nice properties such as invariance under graph isomorphism. For comparing graphs with non-simple spectrums, we propose a procedure to approximate the ELD via a simple perturbation technique to resolve any ambiguity from basis symmetries. We show that such perturbations are stable using matrix perturbation theory under mild assumptions that are straightforward to verify in practice. We demonstrate the excellent applicability of the ELD approach on both simulated and real datasets.
\end{abstract}


\maketitle

\section{Introduction}
Graphs and networks are ubiquitous objects in statistics, machine learning and related fields. A routine task in network analysis is graph comparison, which involves quantifying structural similarities between graphs. Graph comparison is a well-studied problem with an abundant literature. It has been used widely in the biomedical, social and engineering sciences, with application domains ranging from social network analysis to neural connectomics. 

One of the main challenges in graph comparison is to develop simple, fast and theoretically sound methods for comparing graphs that are on different scales, i.e. graphs with different numbers of vertices. This is an important and natural task that arises in many application domains. For example, sociologists are often interested in comparing the structures of social communities with vastly different sizes \citep{grindrod2016comparison}, while biologists might be interested in comparing protein-protein interaction networks between different species \citep{jeong2016effective}. These applications naturally require comparison of graphs with different numbers of vertices. Beyond the scientific applications above, graph comparison can also serve as a foundational tool in many downstream machine learning and statistics tasks, such as graph classification \citep{richiardi2013machine} and statistical modelling of network populations \citep{durante2017nonparametric, lunagomez2020modeling}. Better methods for multiscale graph comparison would extend the reach of many useful graph machine learning algorithms to the multiscale case.

Our motivation is to compare graphs of different sizes based on the similarities in the community structures that they exhibit. One natural mathematical tool well known to capture community structures within a graph is the eigendecomposition of the graph Laplacian. Extensive work in spectral graph theory have shown that the eigenvectors of the graph Laplacian encode important connectivity information within a graph \citep{lee2014multiway, louis2012many}. In particular, the Laplacian eigenvectors specify a Euclidean embedding of the graph's nodes, which forms the foundation of popular algorithms such as spectral clustering \citep{ng2001spectral, shi2000normalized, von2007tutorial}. It is therefore natural to compare the community structures of different graphs using their Laplacian eigenvectors as a starting point. 

Several challenges arise when trying to adopt Laplacian eigenvectors for comparing graphs with potentially different sizes. First, the dimensions of the eigenvectors of each graph would be different and thus not straightforward to compare. Second, ambiguities in the eigenvectors (due to sign-flip or basis symmetries) may arise, leading to difficulty in specifying a well-defined notion of comparison/distance, particularly when the graph has repeated Laplacian eigenvalues. In part due to these challenges, Laplacian eigenvectors/embeddings, while natural, have not found wide-spread application in (multiscale) graph comparison. 

In this work we propose the embedded Laplacian discrepancy (ELD), which aims to bridge these gaps. The ELD provides a simple, fast and natural alternative to existing approaches for graph comparison. The main idea behind ELD is to represent graphs as point clouds in a common, low-dimensional Euclidean space via Laplacian embeddings. This turns the problem of comparing of eigenvectors of different dimensions, which is difficult, into the problem of comparing points clouds in a common Euclidean space, which can be done in a straightforward manner using tools from optimal transport. This leads to an effective measure of dissimilarity. The dimension of the common embedding space, $k$, can be thought of as a "resolution" parameter that dictates the number of communities that one wants to compare \citep{lee2014multiway, louis2012many}. We give practical suggestions on how to choose the hyperparameter $k$ in a principled manner in subsection 3.3.   

A well-known challenge in using eigenvectors for analyzing graphs and performing other machine learning tasks (e.g. principal components analysis) is the potential ambiguity that could arise due to sign-flips and basis symmetries \cite{lim2022sign, eastment1982cross, rustamov2007laplace, bro2008resolving}. Given any eigenvector $\mathbf{v}$, $-\mathbf{v}$ is also a valid eigenvector. When specifying a set of $k$ linearly independent eigenvectors, sign-ambiguity leads to $2^k$ possible configurations. The ELD resolves this ambiguity by symmetrizing the embedding encoded by each Laplacian eigenvector, leading to a well-defined function that is invariant to sign configurations of the Laplacian eigenvectors. 

Ambiguities due to basis symmetries arise when the graphs under consideration have repeated eigenvalues, since the eigenspace under consideration can have infinitely many choices of basis vectors. To tackle ambiguities due to basis symmetries, we break our analysis down into two scenarios. In the case where the graphs under comparison have simple spectrums (i.e. each graph has no repeated eigenvalues), no basis symmetries arise and we show that the ELD is a well-defined pseudometric that is invariant under graph isomorphism. This indicates that ELD compares graphical structure only and is agnostic to permutations of vertex labelings. In the case where there are repeated eigenvalues in the graphs, we propose an approximate procedure (called approximate ELD) that perturbs the Laplacian matrix of the graph with a small, symmetric noise term which will split the repeated eigenvalues almost surely. We then compute the ELD on the perturbed graphs as a proxy for the comparison between the original graphs. We utilize results from matrix perturbation theory that give stability guarantees of such perturbations, under mild conditions that are easily checkable in practice. As such, we bypass the difficult basis-symmetry ambiguity problem by paying the small price of (controllable) perturbation. We demonstrate the practical appeal of the ELD via simulations and experiments. 

Our contributions are: 1) The embedded Laplacian discrepancy is, to the best of our knowledge, the first method that combines a spectral/Laplacian embedding and a Wasserstein approach for comparing \emph{collections} of graphs of different sizes 2) We bypass the ambiguity issues due to sign flips and basis symmetries without resorting to costly intermediate optimization/averaging procedures common in the literature  3) We provide strong evidence, both theoretical and empirical, that supports the efficacy of this novel approach. 

\subsection{Related Work} 
There is an immense literature on graph comparison. Due to space limitations we only highlight those methods that are most relevant to our paper, and delegate the interested reader to the excellent reviews \citep{donnat2018tracking, emmert2016fifty, wills2020metrics, tantardini2019comparing, soundarajan2014guide} for further information. 

A predominant approach for graph comparison is via defining a pairwise graph distance. Distances that are based on the graph spectrum \citep{wilson2008study} or matrix representations \citep{wills2020metrics, koutra2013deltacon} are generally confined to comparing graphs of the same sizes. Feature-based distances, for example those based on spectral or topological descriptors \citep{kaiser2011tutorial, dehmer2010novel, berlingerio2012netsimile}, are difficult to generalize since they operate on a specified list of features that are often hand-picked. The popular graph edit distances \citep{sanfeliu1983distance, gao2010survey, bunke1997relation}, which are extensions of traditional string-based edit distances, have been shown to be NP-hard in general to compute \citep{zeng2009comparing}.

Another approach for comparing graphs of different sizes is by considering graphs as metric spaces and then leveraging distances between isometry classes of metric spaces \citep{wills2020metrics}, e.g. the Gromov-Hausdorff distance \citep{gromov2007metric, gromov1981groups, edwards1975structure} and its relaxations such as the Gromov-Wasserstein distance \citep{chowdhury2021generalized}. The Gromov-Hausdorff distance involves optimizing over isometric embeddings of metric spaces and is known to be NP-hard to compute \citep{chazal2009gromov, memoli2007use, oles2019efficient}. A variety of methods using Wasserstein distances \citep{villani2009optimal} etc. have been proposed for comparing graphs \citep{xu2019gromov, maretic2019got, vayer2018optimal}, although scalability remains a core issue. \citep{lai2017multiscale} compared point clouds in 3D Euclidean space utilizing a sliced Wasserstein approach on the eigensystem of the Laplace-Beltrami operator. 

Beyond graph distances, another predominant approach for graph comparison is graph kernels, which define an inner product between graphs, thus allowing use of kernel-based machine learning methods. \citet{vishwanathan2010graph} provide a unified framework for many graph kernel methods, and \citet{borgwardt2020graph, kriege2020survey, nikolentzos2019graph, ghosh2018journey} provide recent surveys in this area. There are many variants with different emphasis, such as the random walk kernel, graphlet kernels, and shortest path kernels, to name a few. In the multiscale case, the direct product kernel \citep{gartner2003graph, ketkar2009faster} is often used. Given two graphs with $m$ and $n$ vertices respectively, this method constructs a direct product graph of size $O(mn)$, on which computation of the kernel could cost $O((mn)^3)$.  

Some existing methods that address similar problems to ours are outlined below. In the multiscale graph comparison context, one of the most recent methods is the Network Portrait Divergence (NPD) \citep{bagrow2019information}, which uses information theoretic approaches to compare network portraits and is computationally fast. \citep{tsitsulin2018netlsd} proposes NetLSD, a descriptor that utilizes the trace of the Laplacian heat kernel matrix as a summary for graph comparison. We compare the ELD against the NPD and NetLSD in our experiments.

Generally, methods that are based on global graph structure, such as the graph spectrum, do not directly extend to the multiscale scenario. Methods that are based on more local, combinatorial structures, such as those that are based on matching subgraphs, graphlets, edit operations etc, usually run into computational bottlenecks, since at the core of these approaches lies the difficult problem of (sub)graph isomorphism.

\subsection{Organization}
The paper is organized as follows. In section 2, we set up the notation and preliminary notions from spectral graph theory and optimal transport that are needed for this paper. In section 3, we formally introduce the ELD and the approximate ELD. In section 4, we provide a theoretical characterization of the ELD. In section 5, we perform experiments on simulated and real graph data to illustrate the practical appeal of this approach. In section 6, we discuss the future directions and implications of this work. 

\section{Preliminaries}
\subsection{Notation and Setup}
Let the triple $G = (V, E, w)$ denote a simple, undirected, weighted graph, where $V$ is the set of vertices, $E \subset V\times V$ represents the edges of the graph, and $w: E \to \mathbb{R}_0^+$ is a weight function that assigns to each edge a non-negative real number. This framework subsumes the unweighted case, where we simply pick $w$ to be a constant function on the edges. We use $n$ to denote the number of vertices of $G$. Without loss of generality, we use the natural numbers to label the vertices, and denote by $E_{ij}$ that edge that connects vertices $i$ and $j$. We use $[n] := \{1, 2, \cdots, n\}$ to denote the positive natural numbers up to $n$. 

The graph's adjacency matrix $\mathbf{A}$ is the matrix with entries $\mathbf{A}_{ij}:= w(E_{ij}) := w_{ij}$ if there is an edge between nodes $i$ and $j$, and $0$ otherwise. The degree matrix $\mathbf{D}$ is the diagonal matrix with $\mathbf{D}_{ii} = \sum_{j = 1}^n \mathbf{A}_{ij}$. The combinatorial Laplacian matrix $\mathbf{L}$ is defined as $\mathbf{L} = \mathbf{D}- \mathbf{A}$. We use subscripts such as $\mathbf{L}_G$ and $\mathbf{A}_G$ to highlight dependency on the graph $G$. We use $\lambda$ and $\mathbf{v}$ to denote Laplacian eigenvalues and eigenvectors respectively. All matrices considered in the paper are symmetric, hence all corresponding eigenvalues are real. In particular, the combinatorial Laplacian matrix is positive-semidefinite, and its eigenvalues are presented in ascending order, so $0 \leq \lambda_i \leq \lambda_j$ for $i < j$. 
In this paper we only consider graphs that are connected. This has essentially no loss of generality since we are mainly interested in studying community structures of graphs, and in the case of disconnected graphs, the community structures are apparent and each connected component can be analyzed separately. 

\subsection{Laplacian Embeddings}
The core mathematical object in this paper is the combinatorial Laplacian matrix $\mathbf{L} = \mathbf{D}- \mathbf{A}$. It is used widely in spectral graph theory and manifold learning for tasks such as clustering and dimensionality reduction. One useful way to understand the graph Laplacian is to look at its eigendecomposition $\mathbf{L} = \sum_{i = 1}^n \lambda_i \mathbf{v}_i \mathbf{v}_i^T$. As outlined in the seminal paper \cite{belkin2003laplacian}, this decomposition provides a natural Laplacian/spectral embedding of a graph's vertices in Euclidean space: pick the first $k$ eigenvectors (corresponding to the $k$ smallest eigenvalues) of $\mathbf{L}$ (where $k \in \mathbb{N}^+$ is a hyperparameter), and use the entries of these eigenvectors as Euclidean coordinates for embedding the vertices. In other words, the $i$\textsuperscript{th} entry of the $k$\textsuperscript{th} eigenvector $\mathbf{v}_k(i)$ provides the $k$\textsuperscript{th} coordinate of vertex $i$. 

There are many ways to motivate and interpret the Laplacian embedding. The most relevant interpretation for our purposes is to see the Laplacian embedding as the solution to a natural optimization problem, due to \citet{belkin2003laplacian}. Let $\mathbf{Y} := [\mathbf{y}_1 \cdots \mathbf{y}_k]$ be a $n \times k$ matrix, where $\mathbf{Y}_{ij}$ represents the $j$\textsuperscript{th} Euclidean coordinate of the $i$\textsuperscript{th} vertex. If we want an embedding where vertices that are "more connected" with each other are embedded closely, a natural objective function to minimize is $ \sum_{(i,j) \in E} w_{ij}||\mathbf{y}(i) - \mathbf{y}(j)||^2 = \text{Tr}(\mathbf{Y}^T \mathbf{L} \mathbf{Y})$, which is simply a weighted least squares minimization problem. We then have the following result:

\begin{theorem}[Optimal Linear Embedding \citep{belkin2003laplacian}]
Given a connected graph $G = (V, E, w)$,  the solution $\mathbf{Y} = [\mathbf{y}_1 \cdots \mathbf{y}_k]$ of the following optimization problem $$\arg\min_{\mathbf{Y}; \; \mathbf{Y}^T\mathbf{Y} = \mathbf{I}} \text{Tr}(\mathbf{Y}^T \mathbf{L} \mathbf{Y})$$
corresponds to the first $k$ Laplacian eigenvectors.
\end{theorem}
Here the constraint $\mathbf{Y}^T\mathbf{Y} = \mathbf{I}$ enforces orthonormality.  The proof's arguments are based on standard spectral graph theory and the Rayleigh-Ritz variational characterization of eigenvalues, and can be found in references such as \cite{belkin2003laplacian}. Optionally, one can ignore the first eigenvector, which will always be constant. 
  
The loss function above is set up in such a way that the Laplacian embedding respects graph connectivity, since vertices that are "more connected" with each other are embedded more closely. To see this more concretely, consider graphs that exhibit community/cluster connectivity patterns. It is well known that the Laplacian embedding capture such cluster structures faithfully. The Laplacian embedding serves as the foundation of popular algorithms, such as spectral clustering \cite{ng2001spectral, von2007tutorial, shi2000normalized}. The intuition is that since "more connected" vertices are embedded closer together, one can just run Euclidean clustering algorithms (e.g. K-means) in the embedding space to recover graph clusters. The fact that Laplacian embeddings capture cluster structures is rigorously justified by results in spectral graph theory known as higher-order cheeger's inequalities \cite{lee2014multiway, louis2012many}. Such results imply that if the vertices of a graph admit $k$ sparse cuts, then the resulting $k$ partitions can be captured by the first $k$ Laplacian eigenvectors of the graph. 

In summary, if we want to compare graphs based on their connectivity patterns (in particular whether the graphs exhibit similar clustering structures), it is natural to consider the Laplacian embedding as a starting point. The embedding dimension $k$ will serve as a hyperparameter that controls the number of clusters we are taking into consideration. We give suggestions on how to choose $k$ later in the paper. 


\subsection{Wasserstein Distances}

Wasserstein distances define metrics between probability measures. They are intimately connected to optimal transport \citep{villani2009optimal}, and have been used in graph settings increasingly often \cite{saad2021graph, kolouri2020wasserstein}. 

Given two measure $\mu$ and $\nu$ on a $k$-dimensional Euclidean space, define the Wasserstein p-distance between them as:
$$\mathcal{W}_p(\mu, \nu) := \inf_{J \in \mathcal{J}(\mu, \nu)} \left( \int_{\mathbb{R}^k \times \mathbb{R}^k} d(x,y)^p \mathrm{d}J(x,y)\right)^{1/p}$$
where $\mathcal{J}(\mu, \nu)$ denotes the set of all couplings of $\mu$ and $\nu$, and $p \geq 1$ specifies the order of the moment used. 

When $p = 1$, then $\mathcal{W}_1$ is also known as the earth mover distance, and admits the following elegant equivalent representation: 
$$\mathcal{W}_1(\mu, \nu) = \sup_{f \in \mathcal{F}} |\int f(x) d\mu(x) - \int f(y) d\nu(y)| $$
where $\mathcal{F} := \{f: \mathbb{R}^k \to \mathbb{R} \; \given \; |f(x) - f(y)| \leq ||x - y||_2  \; \forall x, y \in \mathbb{R}^k \}$ denotes the set of all Lipschitz functions from $\mathbb{R}^k$ to $\mathbb{R}$. Due to this useful representation, in the rest of the paper we set $p = 1$ for all Wasserstein distances. 

The Wasserstein distance can be easily adapted to compare two sets of points in Euclidean space, by simply specifying $\mu$ and $\nu$ to be the empirical measures of the two sets of points that one wants to compare. 

For cases where the dimension $k > 1$, computing the exact Wasserstein distance can involve time-consuming optimization that hampers computational performance. This computational problem disappears in the case when comparing 1-dimensional distributions. In particular, if $\mu$ and $\nu$ are both measures on the one-dimensional Euclidean space, we have the following simple characterization:
$$\mathcal{W}_1(\mu, \nu) = \int_{0}^1 |F_{\mu}^{-1}(z) -  F_{\nu}^{-1}(z)| dz$$
where $F_{\mu}$ and $F_{\nu}$ are the cumulative distribution functions of $\mu$ and $\nu$ respectively. This formulation allows for one-dimensional Wasserstein distances to be computed very efficiently. The above facts and formulations can be found in standard optimal transport references, such as \citep{kolouri2017optimal, villani2009optimal}.

This motivates us to find principled ways to leverage the computational efficiency of 1-dimensional Wasserstein distances in defining the ELD. In the next section, we utilize the Laplacian embedding and Wasserstein distances in a specific way that is simple and efficient to define the Embedded Laplacian Dicrepancy.

\section{The Embedded Laplacian Discrepancy (ELD)}

We now formally introduce the ELD. We divide our definition and analysis into two parts: 1. comparing graphs with simple spectrums (no repeated eigenvalues) and 2. comparing graphs with non-simple spectrums (with repeated eigenvalues). 

\subsection{The Embedded Laplacian Discrepancy for Graphs with Simple Spectrums}

Let $\mathcal{G}$ denote the space of all simple, undirected, weighted, connected graphs with finite numbers of vertices and simple spectrums. Our goal is to construct a mapping $\rho: \mathcal{G} \times \mathcal{G} \to \mathbb{R}^+_0$ that quantifies the structural similarity/dissimilarity between two graphs. 

We will do so via the following recipe. 1. Compute the top $k$ Laplacian eigenvectors of the two graphs under comparison. 2. Use the entries of the $k$ Laplacian eigenvectors as well as their flipped counterpart (for symmetrization) to represent the nodes of the two graphs as two point clouds in a common $k$-dimensional Euclidean space. 3. Compute the 1-dimensional Wasserstein distance of the point clouds along each of the $k$ canonical Euclidean axes and average them.  

We setup some notation here. Given the $r$th eigenvector $\mathbf{v}^G_r$ of a graph $G$ with $n$ nodes, we associate with it a one-dimensional empirical measure $$\mu_r^G := \frac{1}{2n}\sum_{i = 1}^{n} [\delta(\lambda_r^{G}\mathbf{v}^G_r(i)) + \delta(-\lambda_r^{G}\mathbf{v}^G_r(i))]$$ where $\delta$ is the Dirac measure and $\lambda_r^G$ is the $r$th eigenvalue. Note that the negative of the eigenvector is also included to symmetrize the embedding. In section 4, we show that this leads to embeddings that are invariant to sign-flips of eigenvectors. Notationally, we will use both $\mu$ and $\nu$ below to denote the empirical measures associated with the eigenvectors of different graphs, with $\nu$ defined analogously to $\mu$. 

\begin{definition}[Embedded Laplacian Discrepancy]
Consider two graphs $G_1 = (V_1, E_1, w_1) \in \mathcal{G}$ and $G_2 = (V_2, E_2, w_2)\in \mathcal{G}$, with sizes $n_1:=|V_1|$ and $n_2:=|V_2|$ and Laplacians $L_{G_1}$ and $L_{G_2}$ respectively. Without loss of generality we assume $n_1 \leq n_2$. 
Given a dimension hyperparameter $k \leq n_1$, define the embedded Laplacian discrepancy as
$$\rho(G_1, G_2) := \frac{1}{k}\sum_{r = 1}^k \mathcal{W}_1(\mu_r^{G_1}, \nu_r^{G_2})$$
\end{definition}

Several remarks are in order. First, Laplacian embeddings have traditionally been used in the context of analyzing vertices/clusters in a single graph. One of the unique features of the ELD is it utilizes Laplacian embeddings to project \emph{collections} of graphs with different sizes onto a \emph{common} Euclidean space. Second, we weigh each point inside empirical measures by the corresponding Laplacian eigenvalues, which puts the entries of the Laplacian eigenvectors under their natural scalings. Third, a particular design choice that we have made is to compute the one-dimensional Wasserstein distance along each eigenvector/canonical Euclidean axis. Intuitively, given the orthogonality of a graph's Laplacian eigenvectors, it is natural to compare the distributions along each canonical axis. Alternatively, from the spectral graph theory and graph signal processing literature \citep{Spielman2019Algebraic, shuman2016vertex, ricaud2019fourier}, each Laplacian eigenvector can be thought of as encoding a certain ``frequency" of the graph, and our approach then amounts to comparing the ordered frequency components of each graph one by one. This approach allows us to leverage the efficient computation of one-dimensional Wasserstein distances and directly leads to the ELD being isomorphism invariant (see section 4). 

\subsection{Approximate Embedded Laplacian Discrepancy}

Given two graphs $G_1$ and $G_2$ that we want to compare, when repeated eigenvalues are present in $G_1$ and/or $G_2$, the ELD as defined in the previous subsection is not well-specified. This is due to the basis symmetries that arise from repeated eigenvalues, leading to an infinite number of potential choices of basis vectors that span the eigenspace and ambiguity over which basis to use for the ELD. 

Specifically, if $\lambda$ is an eigenvalue with multiplicity larger than $1$, and $\mathbf{v}_1$ and $\mathbf{v}_2$ are two linearly independent eigenvectors with eigenvalue $\lambda$, then any linear combination of $\mathbf{v}_1$ and $\mathbf{v}_2$ will also be an eigenvector with eigenvalue $\lambda$.  Prior work in the literature \citep{lai2017multiscale} often uses intermediate optimization procedures over the symmetry group to overcome the issue, which could potentially be computationally expensive. Here, we propose an approximate procedure that we term approximate ELD that bypasses the need for any intermediate optimization step. 

Given $G$ as a connected graph with repeated Laplacian eigenvalues. The idea is to add a hollow, symmetric noise matrix to the adjacency matrix of $G$, where each upper-diagonal entry of the noise matrix is the absolute value of an independent Gaussian random variable with variance $\epsilon$. As we demonstrate in section 4, this splits any repeated Laplacian eigenvalues almost surely. We then compute the ELD on these perturbed graphs as a proxy for the comparison between the original graphs. The definition of ELD only depends on the Laplacian eigenvalues and eigenvectors of $G$, which are both stable quantities under perturbation (see section 4) under mild spectral gap conditions that are easily veriable in practice. This, combined with the fact that we can pick $\epsilon$ to be arbitrarily small, leads to a natural approximate notion of ELD that avoids any intermediate optimization step, retains the computational efficiency, and works excellently in practice (see section 5). 

The price that we pay is that the approximate ELD leads to a random quantity, rather than a deterministic one. In practice, one can average over repeated samplings of approximate ELD to obtain mean estimates to the desired level of statistical accuracy.    

\subsection{Selection of hyperparameter $k$}
Both the ELD and the approximate ELD require the user to select a hyperparameter $k$, which represents the number of eigenvectors to consider for both graphs/the dimension of the embedding Euclidean space. A necessary requirement is that when comparing two graphs with $n_1$ and $n_2$ vertices respectively, $k$ is less than or equal to $\min(n_1, n_2)$. 

From the practice of spectral clustering, it is often noted that the existence of a large spectral gap between adjacent Laplacian eigenvalues $\lambda_r$ and $\lambda_{r + 1}$ indicates the "appropriate" number of clusters in the graph is $r$. Recent theoretical work on higher-order Cheeger's inequality (for example Theorem 1.3 in \citep{lee2014multiway}) provides theoretical support for this observation. Hence, for general applications of the ELD, we recommend users to first draw scree plots to inspect the number of meaningful clusters in both graphs, and then pick $k$ accordingly. In general, a larger $k$ provides a finer grain comparison between graphs. When using the approximate ELD, it is important to pick $k$ so that there are no large spectral gaps in the first $k$ eigenvalues of either graph in order for the eigenvectors to be stable. In the case where a repeated eigenvalue is splitted into $b$ different eigenvalues after perturbation, $k$ should be picked in such a way that all $b$ resulting perturbed eigenvectors are included/excluded together. For downstream applications in supervised learning where the ELD is used as an intermediate step, $k$ can be chosen via cross-validation. 

\subsection{Computation of the ELD}

We provide a procedure for implementing the ELD in Algorithm 1. In terms of computational complexity, the dominating step is the eigendecomposition, which is of order $O(n_2^3)$ under a naive theoretical analysis. In practice, most of the time $k << n$ and the eigendecomposition can be sped up via sparse or approximate numerical schemes.  

\begin{algorithm}[]
  \caption{Embedded Laplacian Discrepancy}
  \label{alg:projectedLaplacian}
\begin{algorithmic}[1]
  \STATE \noindent {\bf Input:} Two graphs $G_1 = (V_1, E_1, w_1)$ and $G_2 = (V_2, E_2, w_2)$ with $|V_1| = n_1$ and $|V_2| = n_2$
  \STATE \noindent {\bf Output:} A non-negative real number $\rho(G_1, G_2)$ that quantifies the structural similarity between $G_1$ and $G_2$
  \STATE \noindent {\bf Hyperparameters:} Positive integer $k\leq \min(n_1, n_2)$ 
  \STATE \noindent {\bf Algorithm:}
  \STATE Compute the Laplacians $L_{G_1}$ and $L_{G_2}$ 
  \STATE Compute the eigendecomposition of $L_{G_1}$ and $L_{G_2}$ to obtain the Laplacian eigenvalues/eigenvectors of each graph $\{\lambda^{G_1}_i, \textbf{v}^{G_1}_i\}$ and $\{\lambda^{G_2}_j, \textbf{v}^{G_2}_j\}$ 
  \STATE Construct the $k$ empirical measures for each graph, i.e. construct $\mu_r^{G_1} := \frac{1}{2n_1}\sum_{i = 1}^{n_1} [\delta(\lambda_r^{G_1}\mathbf{v}^{G_1}_r(i)) + \delta(-\lambda_r^{G_2}\mathbf{v}^{G_1}_r(i))]$ and $\nu_r^{G_2} := \frac{1}{2n_2}\sum_{i = 1}^{n_2} [\delta(\lambda_r^{G_2}\mathbf{v}^{G_2}_r(i)) + \delta(-\lambda_r^{G_2}\mathbf{v}^{G_2}_r(i))]$ for each $r$ in $[k]$
    \STATE For each $r \in [k]$, compute the 1D Wasserstein distance $\mathcal{W}_1(\mu_r^{G_1}, \nu_r^{G_2})$
    \STATE Return the average $\rho = \frac{1}{k} \sum_{r = 1}^k \mathcal{W}_1(\mu_r^{G_1}, \nu_r^{G_2})$
    
\end{algorithmic}
\end{algorithm}

\section{Theoretical Properties}

\subsection{Theoretical Properties of ELD}
We now demonstrate some mathematical properties of the ELD in this section. We first show that the ELD is invariant to graph isomorphisms, i.e. permutations of the vertices that preserve edge structure. Given any permutation $\sigma: [n] \to [n]$, we can associate to it a permutation matrix $\mathbf{P}_{\sigma}$, and we use $G_\sigma$ to denote the graph obtained by applying $\sigma$ on the vertices of $G$.

\begin{theorem}[Invariance to Graph Isomorphisms]
The embedded Laplacian discrepancy mapping $\rho: \mathcal{G} \times \mathcal{G} \to \mathbb{R}^+$ is invariant to permutations. I.e. given two graphs $G$ and $H$ with $n_1$ and $n_2$ vertices respectively and simple spectrums, given permutations $\sigma$ on $[n_1]$ and $\omega$ on $[n_2]$, we have:
$$\rho_k(G, H) = \rho_k(G_\sigma, H_\omega)$$
for any integer $k \leq \min(n_1, n_2)$. 
\end{theorem}
\begin{proof}
See section A.2 in the appendix.
\end{proof}

The invariance under isomorphism property ensures that the ELD is only capturing the desired graph connectivity structure, rather than non-structural characteristics of the graphs that depend on arbitrary vertex labels.

We also show that the ELD is invariant to the any sign configuration of the eigenvectors. 
\begin{theorem}[Invariance to Sign Configurations]
The embedded Laplacian discrepancy mapping $\rho: \mathcal{G} \times \mathcal{G} \to \mathbb{R}^+$ is invariant to sign configurations of the eigenvectors. 
\end{theorem}
\begin{proof}
See section A.3 in the appendix.
\end{proof}

When we restrict ourselves to compare connected graphs that do not have repeated eigenvalues, the ELD provides a pseudometric on such graphs. 

\begin{theorem}[Pseudometric Property]
The embedded Laplacian discrepancy mapping $\rho: \mathcal{G} \times \mathcal{G} \to \mathbb{R}^+$ is a pseudo-metric. 
\end{theorem}
\begin{proof}
See section A.1 in the appendix.
\end{proof}

For our purposes of multiscale comparison, the pseudometric property is appropriate and it is in general not feasible to upgrade the ELD to a metric. Since the goal of the ELD is to capture graph structure, isomorphic graphs will have a distance of $0$ between each other under the ELD, thus preventing the realization of the identity of indiscernibles property of metrics. Even if we only consider defining a distance between equivalent classes of isomorphic graphs, the fact that the graphs could be of different sizes renders the identity of indiscernibles property generally impossible since the smaller graph has less degrees of freedom than the larger graph.  

\subsection{Theoretical Properties of approximate ELD}
Here we record some relevant theoretical properties of the approximate ELD procedure. We first demonstrate that the perturbation procedure leads to a graph with distinct eigenvalues almost surely, thus resolving any ambiguity from basis symmetries. 

\begin{theorem}[Perturbation Splits Laplacian Eigenvalues Almost Surely]
Given a graph $G$ with $n$ vertices, adjacency matrix $\mathbf{A}$ and Laplacian matrix $\mathbf{L}$. Define the perturbed adjacency matrix $\Tilde{\mathbf{A}}:= \mathbf{A} + \mathbf{N}$, where $\mathbf{N}$ is a hollow, symmetric matrix where $\mathbf{N}_{ij} \sim |\text{Normal}(0, \epsilon)|$ for every $i < j$ and $\mathbf{N}_{ij} = \mathbf{N}_{ji}$. Define the perturbed diagonal matrix $\Tilde{\mathbf{D}}$ as the matrix with $\Tilde{\mathbf{D}}_{ii} = \sum_{j = 1}^n \Tilde{\mathbf{A}}_{ij}$ and $0$ off the diagonals. Then the perturbed Laplacian matrix $\Tilde{\mathbf{L}} := \Tilde{\mathbf{D}}- \Tilde{\mathbf{A}}$ has no repeated eigenvalues almost surely.  
\end{theorem}
\begin{proof}
See section A.4 in the appendix.
\end{proof}

Next, we record two well-known results in matrix perturbation theory that gives support
\begin{theorem}[Weyl's inequality \citep{chen2021spectral, horn2012matrix}]
Given any $n\times n$ symmetric matrix $\mathbf{L}$ and any $n \times n$ symmetric perturbation matrix $\mathbf{N}$, we have that for any $1 \leq i \leq n$,  
$$|\lambda_i(\mathbf{L}+\mathbf{N}) - \lambda_i(\mathbf{L})| \leq ||\mathbf{N}||$$
where $||\cdot||$ denotes the spectral norm. 
\end{theorem}

\begin{theorem}[Davis-Kahan inequality \citep{davis1970rotation, chen2021spectral}]
Given any $n\times n$ symmetric matrix $\mathbf{L}$ where $\lambda_{k + 1} - \lambda_{k} > \delta > 0$. Let $\mathbf{V} := \sum_{r = 1}^k \mathbf{v}_r\mathbf{v}_r^T$ denote the orthogonal projection operator projecting to the eigenspace spanned by $\mathbf{v}_1, \cdots, \mathbf{v}_k$. Let $\mathbf{N}$ be an $n \times n$ symmetric perturbation matrix. Let $\Tilde{\mathbf{L}} := \mathbf{L} + \mathbf{N}$. Let $\Tilde{\mathbf{V}} := \sum_{r = 1}^k \Tilde{\mathbf{v}}_r\Tilde{\mathbf{v}}_r^T$ denote the orthogonal projection operator projecting to the eigenspace spanned by the eigenvectors of $\Tilde{\mathbf{L}}$, $\Tilde{\mathbf{v}}_1, \cdots, \Tilde{\mathbf{v}}_k$. We then have

$$||\Tilde{\mathbf{V}} - \mathbf{V}||_F \leq \frac{\sqrt{2}||\mathbf{N}||_F}{\delta}$$
where $||\cdot||_F$ denotes the Froebenius norm. 
\end{theorem}

Both Weyl's inequality and the Davis-Kahan inequality are standard results whose proofs can be found in references such as \citep{chen2021spectral}. 
Note that the definition of the ELD/approximate ELD depends only on the graphs under consideration only through their eigenvalues and eigenvectors. Weyl's inequality states that eigenvalues are stable under perturbation. The Davis-Kahan inequality states that the eigenspace spanned by the top $k$ eigenvectors is stable under perturbation, provided that the corresponding spectral gap $\delta$ is not too small. In particular, both bounds depend on the norm of the noise matrix $\mathbf{N}$. Since the user has control over the noise parameter $\epsilon$, in practice the norm of $\mathbf{N}$ could be made small. Moreover, in practice the eigengap $\delta$ of each graph under comparison can be inspected via a simple scree plot. Hence, these results together provide guarantees of the stability of the approximate ELD procedure in practice. 

\section{Experiments and Results}

To gauge the efficacy of the ELD approach in practice, we implemented the ELD algorithm in Python 3.8 and tested it on both simulated and real datasets.  All experiments were conducted on a Unix machine with 32GB of RAM and an Apple Silicon M1 Max chip. The software and data to reproduce all experiments in this paper will be made available after anonymous review. Our implementation detects any repeated eigenvalues in the graphs and automatically switches between the exact ELD and the approximate ELD seamlessly. 

We divide our simulations and experiments into several parts: 1. Comparing deterministic graphs via rings of cliques. 2. Comparing random graphs via stochastic blockmodels. 3. Comparing connectome graphs across different organisms 4. Evaluating results in a downstream graph classification task where the ELD was used as a tool in a K-nearest neighbor classifier. 

We consider two main competitors to the ELD for the multiscale graph comparison task: the Network Portrait Divergence (NPD) \cite{bagrow2019information} and the NetLSD \cite{tsitsulin2018netlsd}. These two competitors are selected based on their popularity, computational efficiency, and applicability to the multiscale case. Many other potential competitor methods for graph comparison do not extend to the case of comparing graphs with different sizes. Other popular methods, such as the influential graph edit distance (GED) \citep{sanfeliu1983distance}, are prohibitively slow. We discovered that the computational time for the GED (using an implementation from the popular NetworkX library \citep{hagberg2008exploring}) exceeded our computational budget even for moderate-sized networks, e.g. graphs with only several hundred vertices. In terms of computational speed in practice, the ELD exceeds the GED by orders of magnitude. We therefore only limit our comparisons to NPD and NetLSD.

\paragraph{Simulated Data: Rings of Cliques} Since there is no ground truth notion of similarity between graphs, the most natural way to evaluate the ELD is by applying it to a collection of graphs that exhibit known structural similarities and observing whether the ELD captures these patterns.

Ring of cliques are a family of graphs that consist of cliques that are connected through single edges. They exhibit very clear community structures. 
In figures 1, 2, and 3, we plot heatmaps that represent dissimilarity/discrepancy matrices computed on a collection of ring of cliques graphs using the ELD ($k = 9$), NPD and NetLSD. We use $R(a, b)$ to denote a ring of cliques graph with $a$ cliques, each clique consisting of $b$ vertices. 

In these figures we observe patterns in the ELD matrix that fit our intuitive understanding of graph structures. Clear block structures in the ELD heatmap reveal the pattern where ring of cliques graphs with similar number of clusters are closer together, whereas the ELD grows larger as the difference between the number of cliques in the graphs grows larger. This confirms our understanding that the ELD capture meaningful cluster/community structures in graphs. 

On the other hand, the NPD and NetLSD heatmaps both exhibited no block structures. Both exhibit a highly checkered pattern instead, indicating a potentially different type of graph structure that the NPD and NetLSD are capturing. These results show that the ELD is able to capture meaningful cluster/community structures in graphs that its competitors cannot. 
 
\begin{figure}
    \includegraphics[width=\linewidth]{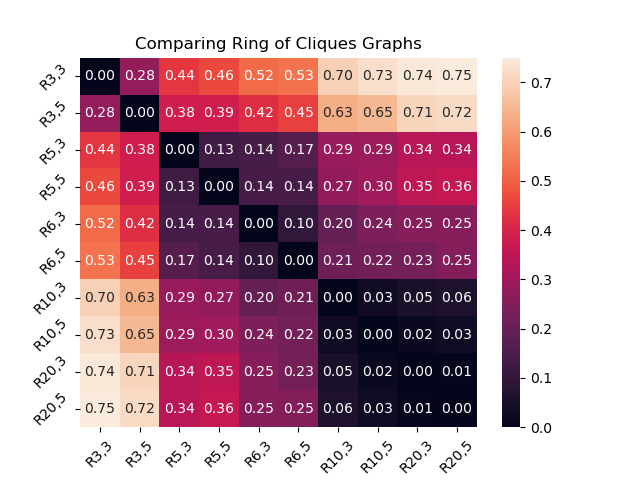}
    \caption{Heatmap of ELD Dissimilarity matrix between rings of cliques}
\end{figure}
\begin{figure}
    \includegraphics[width=\linewidth]{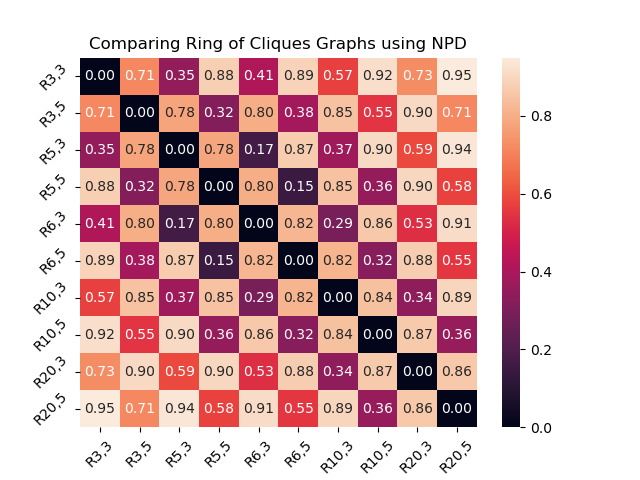}
    \caption{Heatmap of NPD Dissimilarity matrix between rings of cliques}
\end{figure}
\begin{figure}
    \includegraphics[width=\linewidth]{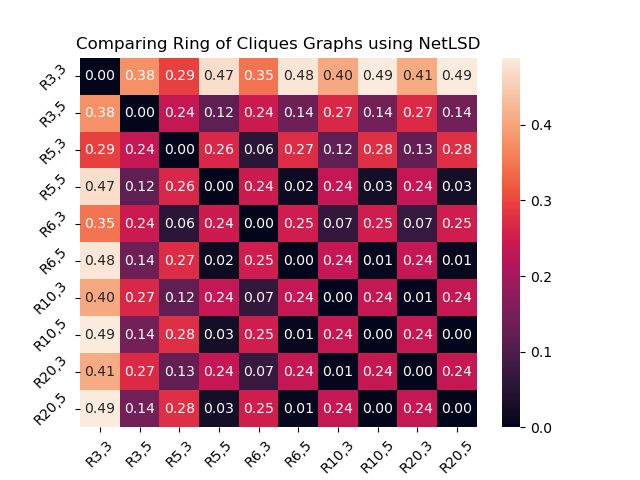}
    \caption{Heatmap of NetLSD Dissimilarity matrix between rings of cliques}
\end{figure}

\paragraph{Simulated Data: Stochastic Blockmodels}
Stochastic blockmodels (SBM) are generative random graph models that naturally carries community structures. 

In figure 4, 5 and 6, we plot heatmaps that represent dissimilarity/discrepancy matrices computed on a collection of stochastic blockmodel graphs using the ELD ($k = 9$), NPD and NetLSD. We use $SBM(a, b)$ to denote an SBM model with $a$ blocks and $b//a$ nodes in each block, with between-block connection probability $0.99$ and within-block connection probability $0.01$. Each entry in the heat map represents the mean value of the discrepancy/dissimilarity measure computed over 20 independent samples of the given SBM models.

\begin{figure}
        \includegraphics[width=\linewidth]{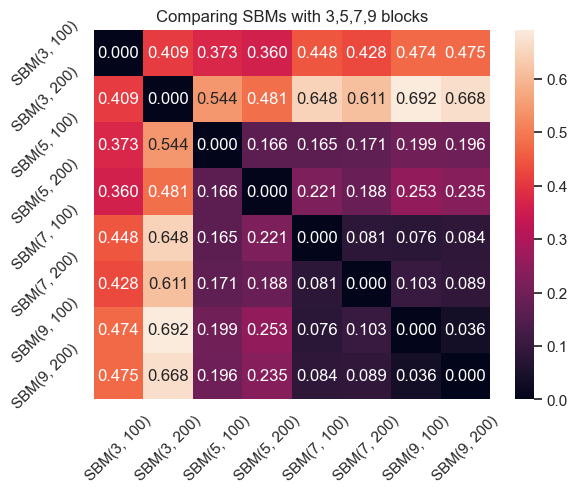}
        \caption{Heatmap of ELD Dissimilarity matrix between SBM graphs}
  \end{figure}
\begin{figure}
        \centering
        \includegraphics[width=\linewidth]{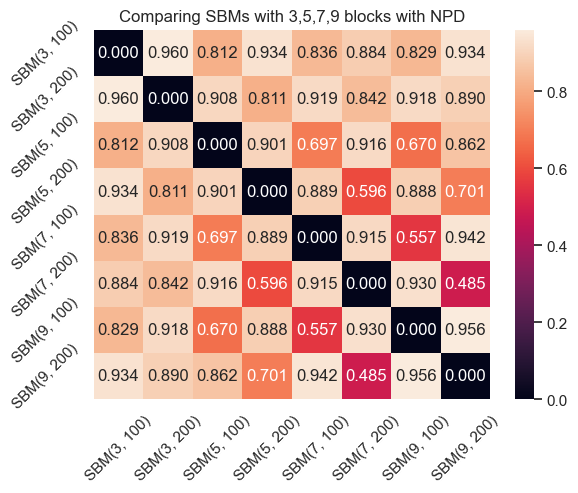}
        \caption{Heatmap of NPD Dissimilarity matrix between SBM graphs}
\end{figure}
 \begin{figure}
        \centering
        \includegraphics[width=\linewidth]{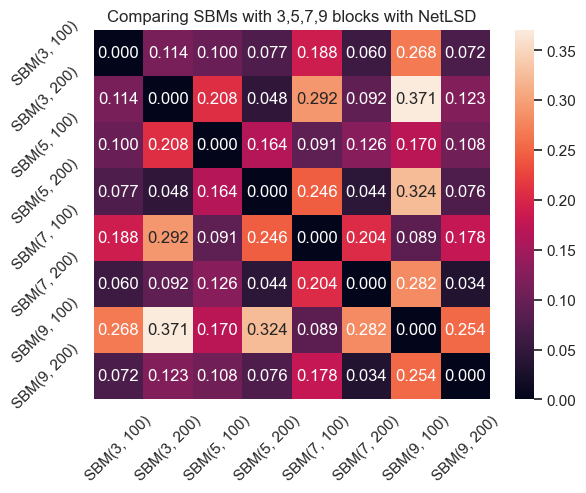}
        \caption{Heatmap of NetLSD Dissimilarity matrix between SBM graphs}
\end{figure}

We observe that, similar to the ring of cliques results, the ELD heatmap exhibited clear and gradual block structures that indicated the capturing of community structures, whereas the NPD and NetLSD heatmaps continue to exhibit a checkered pattern that do not capture community structures. 

\paragraph{Computational Time Comparison}

In figure 7 we compared the computational cost of the ELD, NPD and NetLSD. We used these methods to compare ring of cliques graphs, varying the number of vertices of the graphs compared. To ensure robustness, we repeated these comparisons 20 times and reported the total computer run time (in seconds) of each method. We can observe from figure 7 that all three methods are computationally fast. As the number of vertices increases, all three methods showed a similar growth pattern in run time, with NetLSD being the fastest, ELD being in the middle, and NPD being the slowest. Remark that for graphs of smaller sizes (below 40 vertices) the ELD performs just as fast as NetLSD. 

\begin{figure}
    \includegraphics[width=\linewidth]{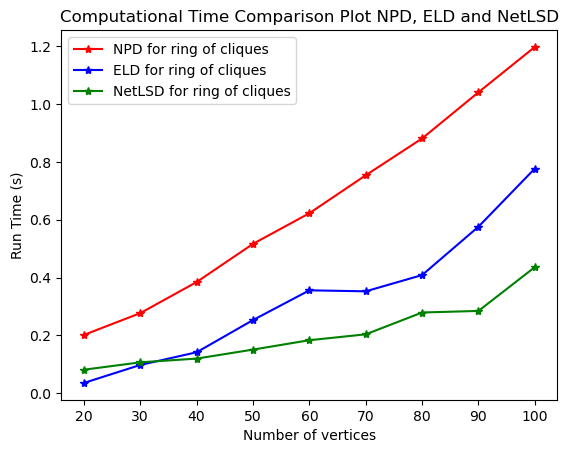}
    \caption{Computational time comparison between ELD, NPD and NetLSD} 
\end{figure}

\paragraph{Real Data: Connectome graphs}
For real network data, we use a connectome dataset hosted at NeuroData \citep{vogelstein2018community}. 

The connectome data we used consisted of graphs characterizing different regions of the brain from different organisms, including two graphs from mice \citep{bock2011network}, three from rats \citep{bota2012combining} and two from the nematode \emph{P. pacificus} \citep{bumbarger2013system}. The connectome graphs are directed, and we take their undirected underlying graphs for comparison in figure 8. 

We observe that the ELD captures relevant graph structural information, since there are clear block structures that distinguishes between the connectome graphs of different organisms in figure 8. 

\begin{figure}
    \includegraphics[width=\linewidth]{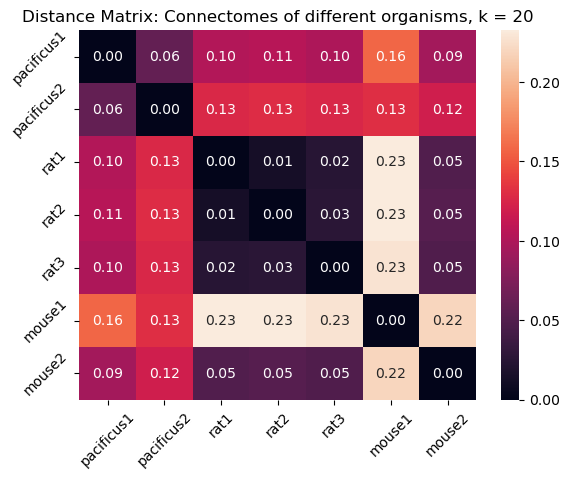}
    \caption{Heatmap of ELD distance matrix between connectomes} 
\end{figure}

\paragraph{Graph Classification: MUTAG dataset}

MUTAG-2 \cite{debnath1991structure} is a dataset on the mutagenicity on Salmonella typhimurium of collection of chemicals know as nitroaromatics. It is a benchmark dataset for graph classificaiton. The dataset consists of 188 chemical graphs, each with a binary label $y$ taking the values $1$ and $-1$ indicating whether the compound is mutagenic.  We obtained the dataset and the dataloader code using the PSCN python package \cite{niepert2016learning}.

We use a $K$-nearest neighbor (KNN) classifier (with $K = 3$) to perform binary classification on the MUTAG dataset, using the ELD, NPD and NetLSD as the dissimilarity/distance measures of the KNN algorithm. We report both training and testing error in table 1. For training error, we trained and evaluated on the entire dataset. For testing error, we trained on 158 data points and tested on 30 hold-out data points. 

Results indicate that these graph distances/discrepancies are capturing meaningful structural information in these chemical graphs that can be leveraged for predictive tasks. In particular, ELD has the best performance overall in terms of both training and testing accuracy. 

\begin{table}
  \caption{Training and Testing Accuracies on MUTAG}
  \label{tab:freq}
  \begin{tabular}{cccl}
    \toprule
    & ELD & NPD & NetLSD\\
    \midrule
   Training Accuracy & \textbf{0.872} & 0.840 & 0.829\\
   Testing Accuracy & \textbf{0.800} & \textbf{0.800} & 0.733\\
   
  \bottomrule
\end{tabular}
\end{table}

\section{Discussion and Future Directions}

The motivation of the ELD is to offer a principled and practical way for comparing graphs that might have similar structures but different sizes, which often arises in applications. Note that comparing graphs of the same size is simply a special case of the ELD framework we presented, so all of our theoretical guarantees and computer programs carry over. The framework that we presented mainly deals with (weighted) simple, undirected graphs, but ELD could be potentially extended to many settings. We discuss two of these directions below. 

\paragraph{Directed Graphs} ELD could potentially be generalized to comparing directed graphs. One way is to apply the ELD approach on the underlying undirected graph of the digraph, but this discards orientation information. Another approach is to take the directed version of the Laplacian \citep{chung2005laplacians} and develop analogous embeddings.  

\paragraph{Labelled Graphs} ELD could also be extended to comparing vertex-labelled graphs, where there are known correspondences between vertices. The most natural way to do this is to retain the vertex correspondences after performing the Laplacian embedding, and compute a distance in Euclidean space between the matching vertices. 

Computational efficiency is one of the advantages of adopting the ELD approach. The most computationally intensive step in ELD is eigendecomposition, which we have implemented using standard numerical libraries and methods. It is possible to achieve substantial speedups in practice by using sparse numerical techniques or a myriad of other approximation/optimization schemes. For applications with repeated calculations on the same graphs (e.g. calculating a distance matrix), the embedded representation can be stored and retrieved for more efficient computation. 

One major design choice made in the definition of the ELD is by using the combinatorial Laplacian rather than the normalized Laplacian $\mathbf{I} - \mathbf{D}^{-1/2}\mathbf{A}\mathbf{D}^{-1/2}$. The ELD definition and implementation described above could easily be extended to the normalized Laplacian case, and virtually all of our theoretical results will carry over. While the combinatorial Laplacian approach naturally takes scale information of the graphs into account, the normalized Laplacian does not. Depending on the context of the application, users might find either forms of the Laplacian to be more appropriate. 

In the experiments, we observed that the NPD/NetLSD dissimilarity matrices did not show block structure, but instead shows an interesting checkerboard structure for comparing rings-of-cliques/SBMs. One hypothesis that this suggests those distances might be capturing local motifs/structural information rather than the number of cliques/communities globally. 

The focus of this paper is mainly to introduce the ELD as a novel discrepancy between graphs of different scales. There are many further downstream machine learning/statistics tasks that one could potentially perform with the ELD, and we delegate these investigations to future research.

\bibliographystyle{ACM-Reference-Format}
\bibliography{sample-base}

\appendix

\section{Appendix}

\subsection{Proof of pseudometric property of ELD}
\begin{proof}
To show that the ELD $\rho_k(\cdot, \cdot)$ is a pseudometric, we have to check four properties. Our arguments hold for all admissible $k$ as specified in the definition of the ELD in section 3.  
\begin{enumerate}
    \item Nonnegativity: $\rho_k(\cdot, \cdot) \geq 0$. Note that $\mathcal{W}_1(\cdot, \cdot) \geq 0$ since it is a metric. Since $\rho_k(G, H)$ is simply an average of $\mathcal{W}_1(\cdot, \cdot)$ terms, $\rho_k$ will also be non-negative. 
    \item Symmetry: $\rho_k(G, H) = \rho(H, G)$ for any graphs $G, H$. Note that since $\mathcal{W}_1(\cdot, \cdot)$ is a metric, it is also symmetric. Since $\rho_k(G, H)$ is just the average of $\mathcal{W}_1(\cdot, \cdot)$ terms, with the first argument of $\mathcal{W}_1$ dependent only on $G$ and the second argument of $\mathcal{W}_1$ dependent only on $H$, conclude that $\rho_k(G, H)$ is also symmetric. 
    \item Identity of indiscernibles: $\rho_k(G, G) = 0$. (since this is a pseudometric, we allow for the possibility that there exist $G \neq H$ where $\rho(G, H) = 0$). To show this, simply note that $\mathcal{W}_1$ is a metric, hence $\mathcal{W}(a, a) = 0$ for any $a = a$. Since $\rho_k(G, G)$ is just the average of $\mathcal{W}_1(\cdot, \cdot)$ terms where the two arguments in each $\mathcal{W}_1$ term will be identical, conclude that each of the $\mathcal{W}_1$ terms will be $0$ and hence $\rho_k(G, G)=0$.
    \item Triangle inequality: $\rho_k(G_1, G_2) + \rho_k(G_2, G_3) \geq \rho_k(G_1, G_3)$ for graphs $G_1, G_2, G_3$. 
    To show the triangle inequality, again leverage the fact that $\mathcal{W}_1$ is a metric to obtain the inequality $\mathcal{W}_1(a, b) + \mathcal{W}_1(b, c) \geq \mathcal{W}_1(a, c)$ for any $a, b, c$.  
    
    Get:
$$\rho(G_1, G_2) + \rho(G_2, G_3)$$ $$= \frac{1}{k}\sum_{r = 1}^{k} \mathcal{W}_1(\mu^{G_1}_r, \mu^{G_2}_r) + \frac{1}{k}\sum_{r = 1}^{k} \mathcal{W}_1(\mu^{G_2}_r, \mu^{G_3}_r) $$

$$= \frac{1}{k}\sum_{r = 1}^{k} \left( \mathcal{W}_1(\mu^{G_1}_r, \mu^{G_2}_r) + \mathcal{W}_1(\mu^{G_2}_r, \mu^{G_3}_r)\right) \geq \frac{1}{k}\sum_{r = 1}^{k} \mathcal{W}_1(\mu^{G_1}_r, \mu^{G_3}_r)$$

$$= \rho(G_1, G_3)$$

\end{enumerate}

\end{proof}

\subsection{Proof of invariance under graph isomorphism}
\begin{proof}
Recall that the ELD between two graphs $G$ and $H$ with $n_1$ and $n_2$ vertices respectively is given by: 
$$\rho_k(G, H) = \frac{1}{k}\sum_{r = 1}^{k} \mathcal{W}_1(\mu^{G}_r, \nu^{H}_r)$$
where $\mu_r^G := \frac{1}{2n_1}\sum_{i = 1}^{n_1} [\delta(\lambda_r^{G}\mathbf{v}^G_r(i)) + \delta(-\lambda_r^{G}\mathbf{v}^G_r(i))]$ and $\nu_r^H := \frac{1}{2n_2}\sum_{j = 1}^{n_2} [\delta(\lambda_r^{H}\mathbf{v}^H_r(j)) + \delta(-\lambda_r^{H}\mathbf{v}^H_r(j))]$.

The ELD by definition computes a distance between the empirical measures associated with the Laplacian embeddings of two respective graphs. 
Hence, to show that the ELD is unaffected by graph isomorphisms, it suffices for us to show that the empirical measure associated with the Laplacian embedding of a graph is invariant under graph isomorphisms/permutaitons. 

Given a graph $G$, a permutation on its vertices $\sigma$, and the corresponding permutation matrix $\mathbf{P}_\sigma$, it is well known fact \citep{Spielman2019Algebraic} that 
$$L_{G_\sigma} = \mathbf{P}_\sigma \mathbf{L}_G\mathbf{P}_\sigma^T$$

It then follows directly by associativity that:
\begin{equation}
    \mathbf{L}_{G_\sigma} \mathbf{v} = \lambda \mathbf{v} \Leftrightarrow (\mathbf{P}_\sigma \mathbf{L}_G\mathbf{P}_\sigma^T)(\mathbf{P}_\sigma\mathbf{v}) = \lambda (\mathbf{P}_\sigma\mathbf{v})
\end{equation}

This shows that the Laplacian eigenvectors of the permuted graph is simply the eigenvectors of the original graph permuted the same way. This also shows that eigenvalues are unchanged by graph isomorphisms/permutations. 

Based on the above, note that the effect of the permutation on the empirical measure of a graph is just changing the order of addition of the Dirac measures, and since we are only considering finite sums, the order of addition does not affect the sum. 

This shows that the empirical measure associated with the Laplacian embedding of a graph is invariant under graph isomorphisms, which proves the desired claim.  
\end{proof}

\subsection{Proof of invariance under sign configurations of eigenvectors}
\begin{proof}
Recall that the ELD between two graphs $G$ and $H$ with $n_1$ and $n_2$ vertices respectively is given by: 
$$\rho_k(G, H) = \frac{1}{k}\sum_{r = 1}^{k} \mathcal{W}_1(\mu^{G}_r, \nu^{H}_r)$$
where $\mu_r^G := \frac{1}{2n_1}\sum_{i = 1}^{n_1} [\delta(\lambda_r^{G}\mathbf{v}^G_r(i)) + \delta(-\lambda_r^{G}\mathbf{v}^G_r(i))]$ and $\nu_r^H := \frac{1}{2n_2}\sum_{j = 1}^{n_2} [\delta(\lambda_r^{H}\mathbf{v}^H_r(j)) + \delta(-\lambda_r^{H}\mathbf{v}^H_r(j))]$.

It suffices that we show the empirical measure of a graph is invariant under sign configurations of eigenvectors. Pick an arbitary eigenvector $\mathbf{v}_r$ from either graph for any $1\leq r \leq k$. Now simply note that if we flip the sign of the chosen eigenvector from $\mathbf{v}_r$ to $-\mathbf{v}_r$, the corresponding empirical measure, as defined, remains unchanged due to the symmetry in the term $\delta(\lambda_r \mathbf{v}_r(j)) + \delta(-\lambda_r\mathbf{v}_r(j))$ for every entry $j$. 
\end{proof}

\subsection{Proof of perturbation almost surely splitting eigenvalues}
\begin{proof}
Note that each entry of the perturbed Laplacian matrix $\Tilde{\mathbf{L}}$ is a random variable following a continuous distribution. It is a well-known mathematical fact that the eigenvalues of a matrix are continuous functions of the entries of the matrix. This is due to the characterization of eigenvalues via the characteristic polynomial (for further reference, see \cite{zedek1965continuity}). Combining the two facts above, conclude that each eigenvalue of $\Tilde{\mathbf{L}}$ is continuously distributed. Conclude that the event where two eigenvalues are exactly equal to each other has measure $0$ under any continuous distribution. 
\end{proof}

\end{document}